\documentclass{article}

\usepackage{xcolor} 
\usepackage{float} 
\usepackage[preprint]{corl_2026} 

\usepackage[utf8]{inputenc}
\usepackage[T1]{fontenc}
\usepackage{hyperref}
\hypersetup{pdftitle={Seeing Through the Illusion: Test-Time Scaling for World Action Models},
  pdfauthor={Cheikh Ahmed, Mahdi Mostajabdaveh, Milad Yazdani, Zirui Zhou}, pdfsubject={}}
\usepackage{url}
\usepackage{xurl} 
\usepackage{booktabs}
\usepackage{amsmath,amssymb,amsfonts}
\usepackage{mathtools}
\usepackage{nicefrac}
\usepackage{microtype}
\usepackage{graphicx}
\usepackage{tikz}
\usetikzlibrary{arrows.meta}
\usepackage{algorithm}
\usepackage{algpseudocode}
\usepackage{enumitem}
\usepackage{cleveref}

\newcommand{\method}{\textsc{WISE}}

\title{Seeing Through the Illusion: Test-Time Scaling for World Action Models}

\author{%
  Cheikh Ahmed$^{1}$ \quad Mahdi Mostajabdaveh$^{1}$ \quad Milad Yazdani$^{2}$ \quad Zirui Zhou$^{1}$\\
  $^{1}$Huawei Technologies Canada\\
  $^{2}$University of British Columbia
}

\begin{document}   

\maketitle

\begin{abstract}
World action models (WAMs) jointly generate actions and imagined futures, but a plausible prediction need not be realized by its co-generated action. We introduce \method{} (\emph{World Action Inference Scaling through Evaluation}), a pre-execution selector that scores each sampled rollout on action validity and temporal regularity, inverse-dynamics consistency between the proposed action and the imagined motion, and instruction-conditioned task progress, and executes the highest-scoring candidate while keeping the WAM frozen. With four candidates per decision, WISE raises closed-loop success from 6.0\% to 24.0\% on RoboLab with Cosmos~3 Edge and from 75.1\% to 78.3\% mean success on LIBERO with DreamZero, and exceeds the best existing selector in aggregate by 4.0 and 0.9 points. Component ablations and agreement analyses show that the three scores rank candidates differently and that task progress carries the largest weight in the fused decision.
\end{abstract}

\section{Introduction}
\label{sec:introduction}

World action models (WAMs) jointly generate robot actions and predictions of how the observed world may evolve~\citep{kim2026cosmospolicy,li2026lingbotva,ye2026dreamzero,wang2026wamsurvey}. This creates a natural opportunity for test-time scaling: sample several complete world-action rollouts from a frozen WAM, evaluate them before execution, and execute the strongest candidate. Sampling and verification have improved action-only policies~\citep{kwok2025robomonkey,zhao2026tapsampling}, while recent WAM systems use predicted value, future consensus, geometric consistency, action-distribution consistency, or action-conditioned simulation as decision-time signals~\citep{kim2026cosmospolicy,ruan2026futurecompatible,zhao2026geobon,zhou2026tau0wm}. These signals judge the imagined future alone or the action alone, or require an external simulator; none checks whether a single candidate's action and future describe the same motion. The central challenge is therefore how to judge a coupled action-future proposal using only information available before physical execution.

\begin{figure}[t]
\centering
\input{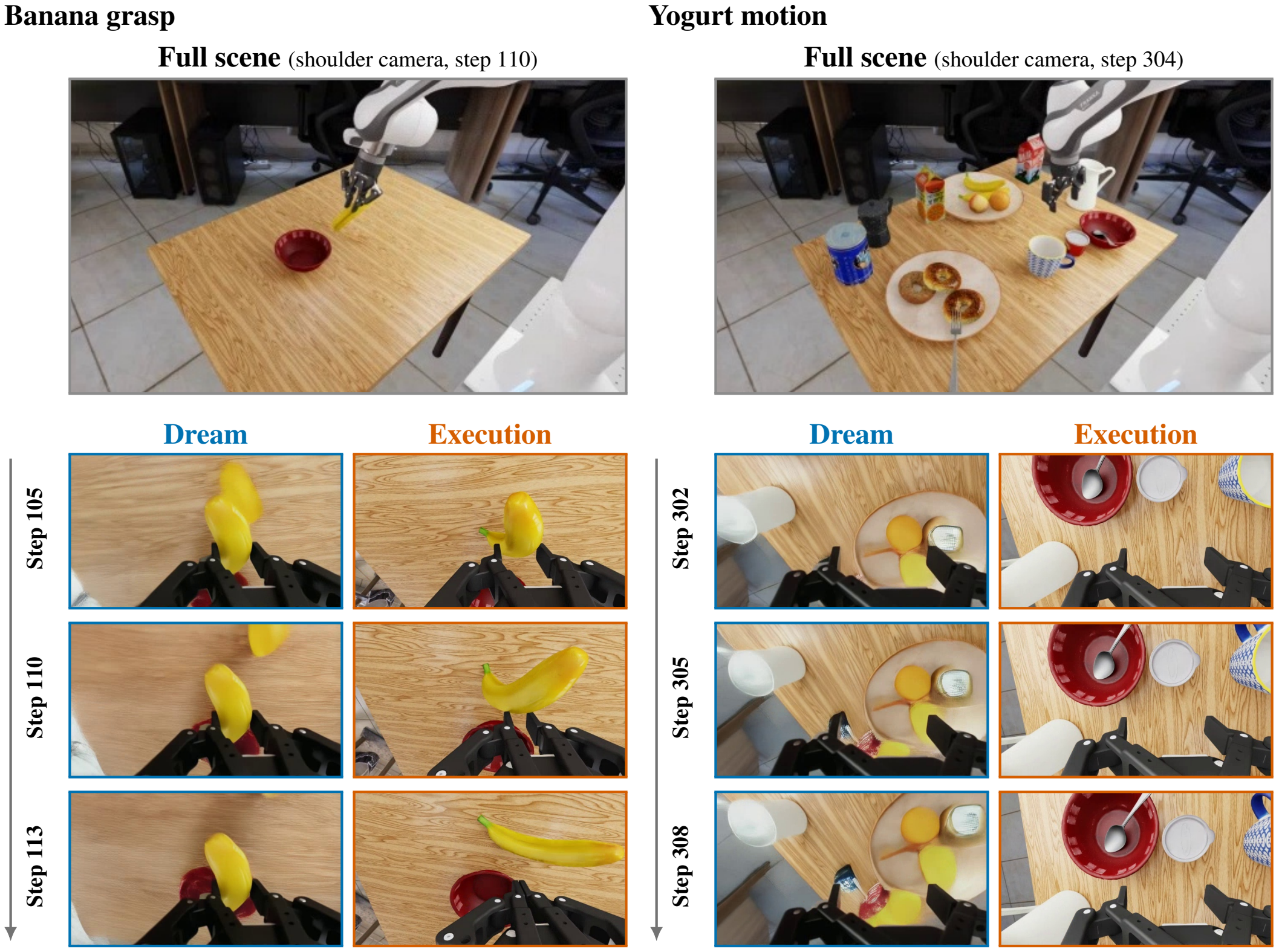}
\caption{Dream-execution divergence for Cosmos~3 Edge on RoboLab. Each pair shows the imagined future (Dream, left) and the realized execution (right) at the indicated step. The banana moves away from the predicted grasp, while the yogurt shows less motion than imagined.}
\label{fig:dream_execution_divergence}
\end{figure}

A WAM's imagined future can diverge from execution, as illustrated in Figure~\ref{fig:dream_execution_divergence}. In the banana-grasp example, the imagined object remains near the gripper while execution displaces it away from the predicted interaction. In the yogurt-motion example, the imagined object moves substantially while execution shows little corresponding motion. Visual plausibility alone does not guarantee that an action will realize the predicted transition. We hypothesize that candidates whose proposed action and imagined future disagree are the ones most likely to diverge in execution, and that this disagreement is measurable before acting.

We introduce \method{}, a pre-execution selector for frozen WAMs. WISE scores each candidate on three properties: whether its action chunk is valid and temporally regular, whether the motion an inverse dynamics model (IDM) recovers from its imagined future agrees with its proposed action, and how much task progress its imagined future represents. The three bounded scores are summed, and the highest-scoring action chunk is executed. The IDM is trained once per embodiment; the WAM is never updated.

We evaluate WISE with Cosmos~3 Edge on RoboLab and DreamZero on LIBERO against Future Consensus, GeoBoN, and Robometer at a matched budget of four candidates. WISE raises success from 6.0\% to 24.0\% on RoboLab and from 75.1\% to 78.3\% on LIBERO, and exceeds the best existing selector in aggregate by 4.0 and 0.9 points.

Our contributions are:
\begin{itemize}[leftmargin=*,itemsep=2pt,topsep=3pt]
  \item A pre-execution selector that combines action validity, cross-output consistency, and task progress in one bounded score, with the WAM frozen.
  \item An inverse-dynamics consistency score that tests whether a candidate's co-generated action supports the motion in its own imagined future.
  \item Closed-loop evaluation on two WAM and embodiment pairs at a matched budget, with component ablations and agreement analyses.
\end{itemize}

\section{Related work}
\label{sec:related_work}

\paragraph{World action models and test-time scaling.}
Recent WAMs couple visual dynamics prediction with action generation, including PAD, Cosmos Policy, DreamZero, LingBot-VA, GR-2, Unified World Models, and VideoVLA~\citep{guo2024pad,kim2026cosmospolicy,ye2026dreamzero,li2026lingbotva,cheang2024gr2,zhu2025uwm,shen2025videovla}. WISE is agnostic to their architecture and training objective; it requires only temporally aligned action and future predictions available before execution. Test-time sampling and verification have improved action-only robot policies~\citep{kwok2025robomonkey,zhao2026tapsampling}. In WAMs, candidate selection has used predicted value, agreement among imagined futures, cross-view geometry, re-denoising consistency, and action-conditioned simulation~\citep{kim2026cosmospolicy,ruan2026futurecompatible,zhao2026geobon,zhou2026tau0wm}. WISE focuses on how to evaluate the native joint rollout itself as a complete control process.

\paragraph{Physical action quality and consistency.}
Robot-action quality has been assessed through smoothness, embodiment constraints, collision/continuity checks, and inverse-dynamics-derived motion rewards~\citep{liu2026evalactions,yang2026vista,wang2026eva,nguyen2026sdn,yang2026physreflect}. These signals operate at different stages: some diagnose completed trajectories, others filter training data or post-train world models, and Selected Diffusion Noise uses a kinematic smoothness criterion directly for inference-time action selection~\citep{nguyen2026sdn}. WISE uses a lightweight version of this idea as one component of its rollout score: it checks the native WAM action against interface constraints and temporal regularity.

Different decision-time methods also use different consistency relations: agreement among alternative futures~\citep{ruan2026futurecompatible}, cross-view geometry~\citep{zhao2026geobon}, compatibility with a learned conditional action distribution~\citep{zhou2026tau0wm}, or recovery of an action supplied to an action-conditioned world model~\citep{seo2026acid}. WISE targets a distinct relation exposed by native WAMs: whether the action and imagined future \emph{co-generated within the same candidate} describe compatible robot motion.

\begin{table}[t]
\centering
\small
\caption{Decision-time consistency criteria compare different quantities.}
\label{tab:consistency_criteria}
\begin{tabular}{p{0.24\linewidth}p{0.46\linewidth}p{0.2\linewidth}}
\toprule
Criterion & Compared quantities & Work \\
\midrule
Future consensus & One predicted future vs. alternative futures & \citep{ruan2026futurecompatible} \\
Cross-view geometry & Multiple views of a predicted future & \citep{zhao2026geobon} \\
Action distribution & Candidate action vs. conditional action distribution & \citep{zhou2026tau0wm} \\
Conditioning fidelity & Input action vs. action recovered from its rollout & \citep{seo2026acid} \\
Cross-output coherence & Co-generated action vs. motion in co-generated future & WISE \\
\bottomrule
\end{tabular}
\end{table}

Table~\ref{tab:consistency_criteria} summarizes the quantities compared by each criterion. In particular, ACID tests whether an IDM recovers the action supplied to an action-conditioned world model~\citep{seo2026acid}. WISE applies this comparison to the action and future co-generated within a native WAM candidate.

\paragraph{Progress and value.}
Task-progress and value models provide a complementary semantic signal. TapSampling uses an action-conditioned progress verifier~\citep{zhao2026tapsampling}; Cosmos Policy and $\tau_0$-WM use predicted value or progress in planning~\citep{kim2026cosmospolicy,zhou2026tau0wm}; Robometer learns frame-level progress and trajectory preferences~\citep{liang2026robometer}; and RISE uses imagined progress values for policy improvement~\citep{yang2026rise}. WISE leaves the task evaluator model-agnostic and uses its frozen judgment as one component of the rollout score.

\section{Problem Formulation}
\label{sec:problem}

We formulate test-time scaling for world action models as an online
selection problem over multiple candidate rollouts. At each control step,
the frozen WAM generates several possible future-action trajectories,
from which a selector chooses one for execution using only pre-execution
information.

At control step $t$, let
$
h_t=(o_t,p_t,\ell)
$
denote the current observation, robot state, and instruction. The frozen
WAM samples $B$ candidate rollouts
\begin{equation}
    c_t^{(i)}
    =
    \left(
        \hat{X}_t^{(i)},
        A_t^{(i)}
    \right)
    \sim
    \pi_\theta(\cdot \mid h_t),
    \qquad
    i=1,\ldots,B,
    \label{eq:candidate}
\end{equation}
where
\[
\hat{X}_t^{(i)}
=
\left(
    \hat{x}_{t+1}^{(i)},
    \ldots,
    \hat{x}_{t+H_v}^{(i)}
\right)
\]
denotes the imagined future rollout and
\[
A_t^{(i)}
=
\left(
    a_t^{(i)},
    \ldots,
    a_{t+H_a-1}^{(i)}
\right)
\]
denotes its co-generated action chunk. Before execution, the candidates
are scored and the highest-ranked rollout is selected:
\begin{equation}
    i_t^*
    =
    \arg\max_{i\in\{1,\ldots,B\}}
    R_{\method}\!\left(c_t^{(i)};h_t\right).
    \label{eq:selection_problem}
\end{equation}
The selected action chunk $A_t^{(i_t^*)}$ is then executed, after which
the robot observes the resulting state and replans. The WAM remains
frozen throughout; test-time scaling acts only through candidate
generation and pre-execution selection.

\section{WISE: Selecting WAM Rollouts}
\label{sec:method}

\subsection{Rollout scoring}
\label{sec:process_reward}

\begin{figure}[t]
\centering
\input{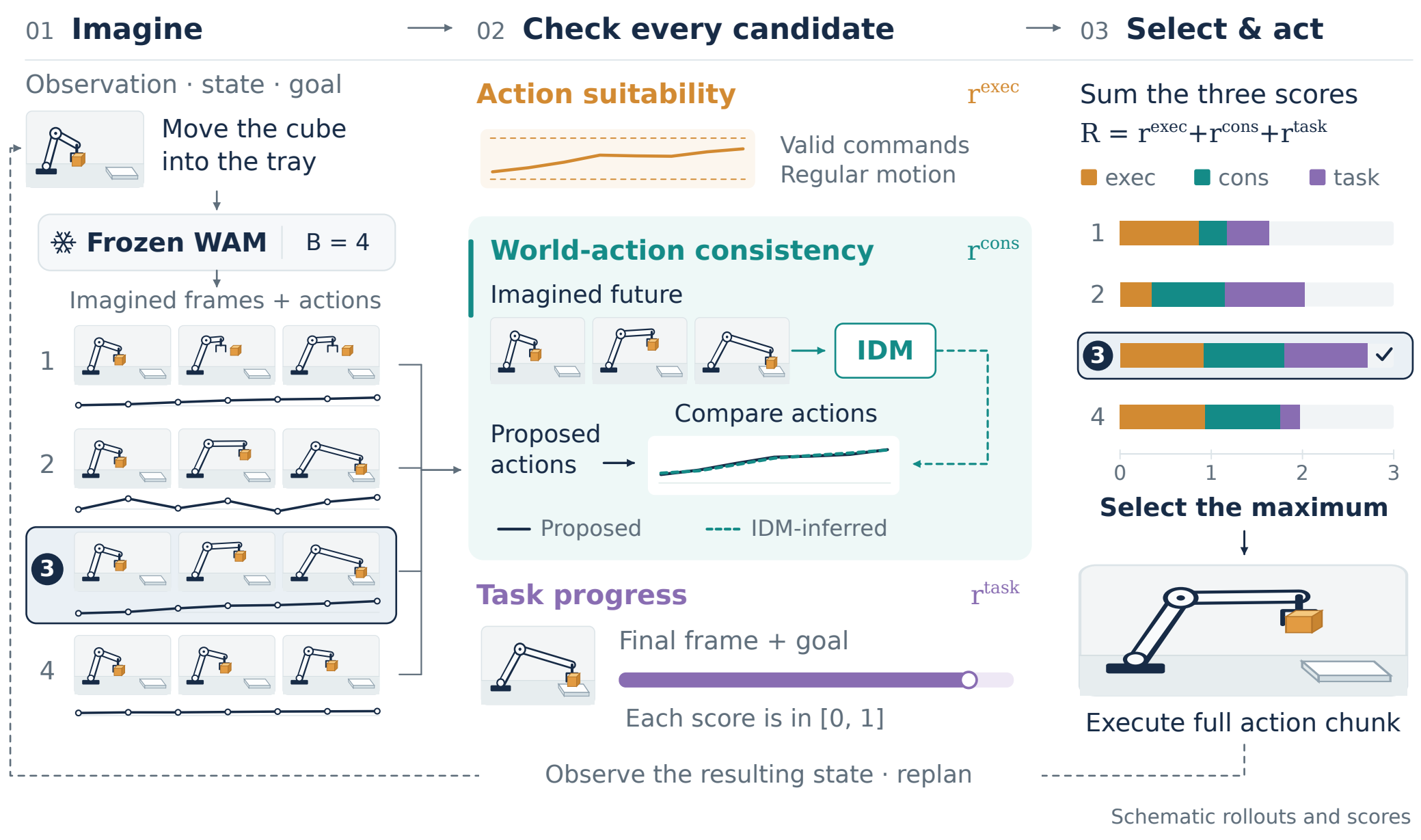}
\caption{WISE at each control decision. A frozen WAM samples $B$ paired
action--future candidates ($B=4$ in our evaluations). Each candidate
receives three scores in $[0,1]$. The IDM maps imagined frames back to
actions for comparison with the candidate's proposed controls; the task
evaluator scores progress at the final imagined frame. WISE sums the
scores, executes the selected action chunk, and observes the resulting
state before replanning.}
\label{fig:wise_pipeline}
\end{figure}

For candidate
$c_t^{(i)}=(\hat{X}_t^{(i)},A_t^{(i)})$,
WISE defines
\begin{equation}
    R_{\method}\!\left(c_t^{(i)};h_t\right)
    =
    r_{t,i}^{\mathrm{exec}}
    +
    r_{t,i}^{\mathrm{cons}}
    +
    r_{t,i}^{\mathrm{task}},
    \label{eq:process_reward}
\end{equation}
where the components score action suitability, agreement between the
co-generated action and imagined motion, and predicted task progress.
They address different potential failures: smooth actions may make no
task progress, and apparently useful futures may be incompatible with
their proposed actions. Figure~\ref{fig:wise_pipeline} summarizes the
scoring and selection procedure.

\subsection{Action validity and temporal regularity}
\label{sec:executability}

The execution component evaluates whether a candidate proposes an action
sequence that is physically plausible to execute. A candidate may predict
a desirable future while containing commands that violate the robot's
operating constraints or imply implausible motion. We therefore assess the
proposed action chunk directly using validity checks defined by the
corresponding control interface.

For candidate $i$, let $u_h^{(i)} \in \mathbb{R}^{d_c}$, $h=0,\ldots,H_a-1$,
denote the continuous control components of the $h$-th action in the
chunk $A_t^{(i)}$ of Equation~\ref{eq:candidate}. Executability is assessed
through complementary validity checks. At the most direct level, commanded
values should remain within their admissible ranges. When the control
representation admits a physical interpretation across time, additional
checks can be derived from consecutive commands.

For example, if $u_h^{(i)}$ is a position-like command sampled at interval
$\Delta t$, the implied velocity, acceleration, and jerk are
\[
    v_h^{(i)} = \frac{u_h^{(i)} - u_{h-1}^{(i)}}{\Delta t},
    \qquad
    \alpha_h^{(i)} = \frac{v_h^{(i)} - v_{h-1}^{(i)}}{\Delta t},
    \qquad
    j_h^{(i)} = \frac{\alpha_h^{(i)} - \alpha_{h-1}^{(i)}}{\Delta t}.
\]
Even when every commanded position lies within its allowed range, the chunk
may still be implausible if consecutive commands imply excessive velocity or
acceleration, so position, velocity, acceleration, and jerk are treated as
complementary checks of the same property.

For each applicable validity check $k$, let
$\delta_k^{(i)} \geq 0$ denote the severity of its worst violation in the
candidate chunk, measured in the units of that check. We map this violation
to a soft score
\begin{equation}
    S_k^{(i)}
    =
    \exp\left(
        -\frac{\delta_k^{(i)}}{\tau_k}
    \right),
\end{equation}
where $\tau_k > 0$ is the tolerance of check $k$ in the same units. Satisfied constraints receive
$S_k^{(i)} = 1$, while increasingly severe violations smoothly reduce the
score toward zero rather than causing a hard rejection.

The execution score combines the checks supported by the control interface:
\begin{equation}
    r_{t,i}^{\mathrm{exec}}
    =
    \frac{
        \sum_{k \in \mathcal{K}}
        w_k S_k^{(i)}
    }{
        \sum_{k \in \mathcal{K}}
        w_k
    }.
    \label{eq:execution_score}
\end{equation}
Higher values indicate action chunks that better respect the robot's physical
and control constraints.
\subsection{World-action consistency}
\label{sec:consistency}

A WAM candidate jointly predicts an action trajectory and an imagined
future. WISE evaluates whether the co-generated action agrees with the
robot motion inferred from that imagined future.

Let $f_\phi$ denote an inverse dynamics model (IDM) trained on
embodiment-specific demonstration trajectories. Applied to the imagined future $\hat{X}_t^{(i)}$, the IDM
predicts a continuous action trajectory
$
    \tilde{A}_{\mathrm{c},t}^{(i)}
    \in
    \mathbb{R}^{H_a\times d_c},
$
together with any discrete control predictions required by the
embodiment. The IDM outputs normalized coordinates
$z_{\mathrm{idm},t+h}^{(i)}$, which are mapped back to the WAM action
space using the IDM training-set statistics:
\begin{equation}
    \tilde{a}_{\mathrm{c},t+h}^{(i)}
    =
    z_{\mathrm{idm},t+h}^{(i)}
    \odot
    \boldsymbol{\sigma}_{\mathrm{train}}
    +
    \boldsymbol{\mu}_{\mathrm{train}}.
    \label{eq:idm_unnormalization}
\end{equation}

\paragraph{Continuous-action consistency.}
Let
$
A_{\mathrm{c},t}^{(i)}
\in
\mathbb{R}^{H_a\times d_c}
$
denote the continuous channels of the WAM-generated action chunk
$A_t^{(i)}$. We compute the mean absolute discrepancy in standardized
training-set units:
\begin{equation}
    L_{t,i}^{\mathrm{cont}}
    =
    \frac{1}{H_a d_c}
    \sum_{h=0}^{H_a-1}
    \sum_{j=1}^{d_c}
    \frac{
        \left|
            \tilde{a}_{\mathrm{c},t+h,j}^{(i)}
            -
            a_{\mathrm{c},t+h,j}^{(i)}
        \right|
    }{
        \sigma_{\mathrm{train}}^{(j)}
    }.
    \label{eq:continuous_consistency_error}
\end{equation}
The discrepancy is converted to a consistency score
\begin{equation}
    r_{t,i}^{\mathrm{cont}}
    =
    \exp\!\left(
        -L_{t,i}^{\mathrm{cont}}
    \right).
    \label{eq:continuous_consistency}
\end{equation}

\paragraph{Discrete-action consistency.}
For discrete control channels such as the gripper, we compare the command
inferred from the imagined rollout with the command proposed by the WAM.
After mapping both predictions to discrete states using a decision threshold,
the consistency score is defined as
\begin{equation}
    r_{t,i}^{\mathrm{disc}}
    =
    \frac{1}{H_a}
    \sum_{h=0}^{H_a-1}
    \mathbb{I}
    \left[
        \mathbb{I}
        \left(
            \hat{g}_{\mathrm{idm},t+h}^{(i)} > \tau_g
        \right)
        =
        \mathbb{I}
        \left(
            g_{\mathrm{wam},t+h}^{(i)} > \tau_g
        \right)
    \right].
    \label{eq:discrete_consistency}
\end{equation}

The final world-action consistency score combines continuous-motion and
discrete-control agreement:
\begin{equation}
    r_{t,i}^{\mathrm{cons}}
    =
    \frac{1}{2}
    r_{t,i}^{\mathrm{cont}}
    +
    \frac{1}{2}
    r_{t,i}^{\mathrm{disc}}.
    \label{eq:consistency_reward}
\end{equation}
A high consistency score indicates that the co-generated action and
imagined future provide compatible descriptions of the candidate robot
motion. This is an internal property of the generated rollout;
correspondence between the imagined future and the subsequently realized
environment remains a separate source of uncertainty.

\subsection{Task progress and rollout selection}
\label{sec:task_reward}

A candidate may satisfy the execution constraints and remain consistent
with its imagined future while still making little progress toward the
instruction. WISE therefore uses a task evaluator $q_\psi$ to estimate the
task progress represented by each imagined rollout.

For candidate $i$, the task score is defined as
\begin{equation}
    r_{t,i}^{\mathrm{task}}
    =
    q_\psi
    \left(
        o_t,
        \hat{X}_t^{(i)},
        \ell
    \right),
    \label{eq:task_reward}
\end{equation}
where the evaluator receives the current observation $o_t$, the imagined
rollout $\hat{X}_t^{(i)}$, and the instruction $\ell$. Higher scores are
assigned to imagined futures that indicate greater progress toward completing
the instructed task.

The scoring framework permits different task evaluators. In both
experimental settings, we use a frozen Robometer model and take the
progress score of the final imagined frame, as specified in
Section~\ref{sec:experimental-setup}.

WISE sums the component scores using Equation~\ref{eq:process_reward}
and selects the maximum using Equation~\ref{eq:selection_problem}.
It then executes the selected action chunk and replans from the next
observation.

\section{Experiments}
\label{sec:experiments}

We evaluate WISE with two frozen WAM-embodiment settings, Cosmos~3 Edge on RoboLab and DreamZero on LIBERO, and compare against existing rollout selectors under a matched candidate budget.

\subsection{Experimental Setup}
\label{sec:experimental-setup}

\paragraph{RoboLab.}
We evaluate Cosmos~3 Edge~\citep{nvidia2026cosmos3edge} on 10
manipulation tasks from RoboLab~\citep{yang2026robolab}. Each task is
evaluated over five closed-loop episodes, yielding 50 episodes in total.

\paragraph{LIBERO.}
We evaluate DreamZero~\citep{ye2026dreamzero} on the four standard
LIBERO suites, LIBERO-Spatial, LIBERO-Object, LIBERO-Goal, and
LIBERO-Long~\citep{liu2023libero}, using the publicly released RLinf
WAN2.2-5B LIBERO SFT checkpoint at training step 26,000~\citep{yu2026rlinf,rlinf2026dreamzero26000}. Each suite contains 10
language-conditioned manipulation tasks, and we evaluate 500 closed-loop
episodes per suite, yielding 2,000 episodes per method across the four
suites.

\paragraph{WISE evaluators.}
For task progress, we use Robometer~\citep{liang2026robometer}:
Robometer-4B on RoboLab and Robometer-LIBERO on LIBERO, with
$r_{t,i}^{\mathrm{task}}\in[0,1]$ given by the final imagined frame's progress
score. For consistency, we train embodiment-specific vision-only IDMs on
DROID~\citep{khazatsky2024droid} real-robot trajectories for RoboLab and
the \texttt{physical-intelligence/libero} demonstration release for LIBERO.
The RoboLab IDM uses 21,000 training and 1,000 scene-disjoint validation
episodes. For LIBERO, we use all four standard suites (1,693 episodes in
total), with 5\% held out for validation. Both IDMs predict normalized continuous controls and gripper
state using Smooth L1 and binary cross-entropy objectives, respectively.
The prediction horizon is matched to each WAM: the RoboLab IDM maps a
33-frame rollout to the corresponding 32-step Cosmos~3 Edge action chunk,
whereas the LIBERO IDM predicts DreamZero's 16-step action chunk from its
9-frame two-view decoded clip.

\paragraph{Evaluation protocol and baselines.}
In both settings, the underlying WAM is kept frozen throughout evaluation. Direct execution uses a single sampled world-action rollout per decision step ($B=1$), whereas WISE and all test-time selectors use $B=4$. We compare WISE with Future Consensus~\citep{ruan2026futurecompatible}, GeoBoN~\citep{zhao2026geobon}, and Robometer~\citep{liang2026robometer}. Each method begins from the task's prescribed initial condition and then runs its own closed-loop trajectory: once selectors choose different action chunks, their subsequent observations and candidate rollouts naturally diverge. After a rollout is selected, its full action chunk is executed before replanning. We report closed-loop task success rate.

\subsection{Main Results}
\label{sec:main-results}

\begin{table*}[h]
    \centering
    \small
    \begin{minipage}[t]{0.40\textwidth}
        \centering
        \caption{Success Rate (\%) on RoboLab.}
        \label{tab:robolab-results}
        \vspace{2pt}
        \begin{tabular}{lc}
            \toprule
            Method & Success (\%) \\
            \midrule
            Cosmos~3 Edge ($B=1$)       & 6.0 \\
            \quad + Future Consensus    & 18.0 \\
            \quad + GeoBoN              & 20.0 \\
            \quad + Robometer           & 20.0 \\
            \quad + \textbf{WISE (Ours)} & \textbf{24.0} \\
            \bottomrule
        \end{tabular}
    \end{minipage}%
    \hfill
    \begin{minipage}[t]{0.60\textwidth}
        \centering
        \setlength{\tabcolsep}{4.0pt}
        \caption{Success Rate (\%) on LIBERO.}
        \label{tab:libero-results}
        \vspace{2pt}
        \begin{tabular}{lccccc}
            \toprule
            Method & Spatial & Object & Goal & Long & Avg. \\
            \midrule
            DreamZero ($B=1$)
                & 92.6 & 88.2 & 51.6 & 67.8 & 75.1 \\
            \quad + Future Consensus
                & 90.6 & 90.0 & 53.4 & 69.6 & 75.9 \\
            \quad + GeoBoN
                & 91.2 & 90.8 & \textbf{55.0} & 72.0 & 77.3 \\
            \quad + Robometer
                & 91.2 & 92.6 & 53.6 & \textbf{72.2} & 77.4 \\
            \quad + \textbf{WISE (Ours)}
                & \textbf{95.8} & \textbf{93.0} & 53.6 & 70.8 & \textbf{78.3} \\
            \bottomrule
        \end{tabular}
    \end{minipage}
\end{table*}
\begin{table*}[t]
    \centering
    \small

    \begin{minipage}[t]{0.61\textwidth}
        \centering
        \caption{Average episode runtime relative to single-rollout execution ($B=1$).}
        \label{tab:runtime}
        \vspace{2pt}
        \setlength{\tabcolsep}{4.5pt}
        \begin{tabular}{lcccc}
            \toprule
            & \multicolumn{2}{c}{RoboLab}
            & \multicolumn{2}{c}{LIBERO} \\
            \cmidrule(lr){2-3}\cmidrule(lr){4-5}
            Method & s/ep. & Rel. & s/ep. & Rel. \\
            \midrule
            $B=1$
                & 892.5   & 1.0$\times$
                & 7.1    & 1.0$\times$ \\
            Future Consensus
                & 2063.1 & 2.3$\times$
                & 20.5   & 2.9$\times$ \\
            GeoBoN
                & 2203.0 & 2.5$\times$
                & 29.6   & 4.2$\times$ \\
            Robometer
                & 2776.3 & 3.1$\times$
                & 25.5   & 3.6$\times$ \\
            \textbf{WISE (Ours)}
                & 3062.5 & 3.4$\times$
                & 25.2   & 3.5$\times$ \\
            \bottomrule
        \end{tabular}
    \end{minipage}%
    \hfill
    \begin{minipage}[t]{0.36\textwidth}
        \centering
        \setlength{\tabcolsep}{3.5pt}
        \caption{RoboLab single-component selector ablation.}
        \label{tab:selector_ablation}
        \vspace{2pt}
        \begin{tabular}{lc}
            \toprule
            Method & Success (\%) \\
            \midrule
            Exec-only             & 10.0 \\
            Cons-only             & 16.0 \\
            Prog-only (Robometer) & 20.0 \\
            \textbf{WISE}         & \textbf{24.0} \\
            \bottomrule
        \end{tabular}
    \end{minipage}

\end{table*}
WISE improves over single-rollout execution in both settings
(Tables~\ref{tab:robolab-results} and~\ref{tab:libero-results}). On
RoboLab, success increases from 6.0\% for the base model to 24.0\%
for WISE. At the same candidate budget, Future Consensus, GeoBoN, and
Robometer obtain 18.0\%, 20.0\%, and 20.0\%. WISE leads the
strongest competing selectors by 4.0 percentage points, or two
additional successes in 50 episodes.

On LIBERO, WISE improves over single-rollout execution of the frozen
DreamZero model on all four suites and raises mean success from
75.1\% to 78.3\%. The largest gains are on LIBERO-Object (+4.8
points) and LIBERO-Spatial (+3.2 points), where WISE is also the
strongest selector. GeoBoN leads on LIBERO-Goal and Robometer on
LIBERO-Long. In aggregate, WISE leads Robometer by 0.9 points, about
one standard error at 2,000 episodes per method.

Table~\ref{tab:runtime} reports the corresponding wall-clock cost.
Relative to single-rollout execution, WISE increases average episode
runtime by $3.4\times$ on RoboLab and $3.5\times$ on LIBERO. Among
the $B=4$ selectors, WISE has the highest runtime on RoboLab, while on
LIBERO its cost is comparable to Robometer and lower than GeoBoN.
\subsection{Ablation Analysis}
\label{sec:ablation}

We analyze the role of the individual WISE scoring components from three
complementary perspectives. We first evaluate separate closed-loop runs in
which rollout selection is restricted to a single component, providing direct
evidence of how single-component selection affects task performance. We then
analyze the per-candidate scores logged during the full WISE evaluations to
study how the components agree with the fused selector and how removing each
component changes its decisions.
\subsubsection{Single-Component Selector Ablation}
\label{sec:selector_ablation}

We first evaluate single-component variants under the same experimental setup
and closed-loop protocol described in Section~\ref{sec:experimental-setup}.
The only change is the rollout-selection criterion: Exec-only selects using
the execution score, Cons-only selects using the consistency score, and
Prog-only selects using the task-progress score.

As shown in Table~\ref{tab:selector_ablation}, each single-component
selector trails the combined score: Exec-only achieves 10.0\%, Cons-only
16.0\%, and Prog-only (Robometer) 20.0\%, compared with 24.0\% for
WISE.

\subsubsection{Component Agreement}
\label{sec:component_agreement}

We next examine whether the execution, consistency, and task-progress scores
independently favor the same candidate as the fused WISE selector. At each
best-of-$B$ decision, we identify the candidate preferred by each component
and compare it with the candidate selected by WISE. We additionally measure
how often all three components independently favor the same candidate.

As shown in Table~\ref{tab:component_agreement}, task progress has the highest
agreement with WISE in both settings, yet differs from the fused choice in
roughly one third of decisions. All three components favor the same
candidate in 7.8\% of RoboLab decisions and 7.0\% of LIBERO decisions,
so the three scores rank candidates differently in more than 92\% of
decisions.

\subsubsection{Counterfactual Component Removal}
\label{sec:leave_one_out}

Finally, we use the same full-run logs to counterfactually recompute the WISE
selection after removing one scoring component at a time. We report the
fraction of decisions for which removing a component changes the selected
candidate.

Table~\ref{tab:leave_one_out} shows that task progress has the largest effect
on the fused decision: removing it changes nearly half of the selections in
both settings. The remaining components also influence the final choice, but
their relative effect differs across environments. Execution has a larger
effect in RoboLab, whereas consistency has a larger effect in LIBERO,
suggesting that the contribution of each signal depends on the candidate
distribution and evaluation setting. These analyses describe how each
component shapes the selection; the effect on closed-loop success is
measured by the single-component runs in Table~\ref{tab:selector_ablation}.

\begin{table*}[t]
    \centering
    \small

    \begin{minipage}[t]{0.49\textwidth}
        \centering
        \caption{Agreement between individual scoring components and the
        final WISE rollout selection.}
        \label{tab:component_agreement}
        \vspace{2pt}
        \begin{tabular}{lcc}
            \toprule
            Agrees with WISE & RoboLab & LIBERO \\
            \midrule
            Execution     & 44.4\% & 34.5\% \\
            Consistency   & 36.9\% & 48.4\% \\
            Task progress & 67.2\% & 65.5\% \\
            All three     & 7.8\%  & 7.0\%  \\
            \bottomrule
        \end{tabular}
    \end{minipage}%
    \hfill
    \begin{minipage}[t]{0.49\textwidth}
        \centering
        \caption{Fraction of WISE selections that change when each scoring
        component is removed.}
        \label{tab:leave_one_out}
        \vspace{2pt}
        \begin{tabular}{lcc}
            \toprule
            Component removed & RoboLab & LIBERO \\
            \midrule
            Execution     & 24.7\% & 12.4\% \\
            Consistency   & 16.8\% & 30.4\% \\
            Task progress & 47.1\% & 48.5\% \\
            \bottomrule
        \end{tabular}
    \end{minipage}

\end{table*}
\subsection{Discussion}
At the shared candidate budget of $B=4$, WISE is the strongest selector in aggregate in both settings. Its margin over the strongest competing selector is smaller than its gain over single-rollout execution, and GeoBoN and Robometer lead on two LIBERO suites.

\paragraph{Limitations.}
WISE depends on its auxiliary evaluators: the IDM may misrecover the actions implied by an imagined trajectory, and the progress model may misjudge visually plausible futures. The single-component runs measure each score on its own; they do not measure the incremental benefit of each score in the presence of the other two. The RoboLab evaluation uses 50 episodes per method, so margins of a few points between selectors are within sampling variation. Test-time selection multiplies inference cost by the number of candidates, and WISE adds the IDM and progress-model evaluations on top.

\paragraph{Beyond test-time selection.}
Although WISE is used here for test-time selection, its rollout score could support broader uses. At inference time, it could enable adaptive replanning, where low-scoring candidates trigger resampling or earlier replanning rather than executing the full action chunk. Beyond inference, the same score could serve as a reward for WAM post-training, encouraging executable, world-action-consistent, and task-directed generations. We leave this direction to future work.

\section{Conclusion}
\label{sec:conclusion}
We introduced \method{}, a selector that scores paired action--future rollouts from frozen WAMs before execution on action validity, inverse-dynamics consistency, and task progress. With Cosmos~3 Edge on RoboLab and DreamZero on LIBERO, WISE raises closed-loop success over direct execution and is the strongest selector in aggregate at a matched budget of four candidates. Future work includes seeded evaluation at larger episode counts, pairwise component ablations, and using the WISE score as a post-training reward.

\clearpage
\bibliography{references}



\end{document}